\documentclass[secnumarabic, graphics,floatfix, nofootinbib,tightenlines,nobibnotes, aps, prx, twocolumn]{revtex4-1}
\usepackage{amssymb}
\usepackage{amsmath}               
  {
      \newtheorem{assumption}{Assumption}
      \newtheorem{theorem}{Theorem}
      \newtheorem{definition}{Definition}
      
      \newtheorem{property}{Property}
  }
\usepackage{graphicx}
\usepackage{dcolumn}
\usepackage{bm}
\usepackage{braket}
\usepackage{subfigure}
\usepackage{enumerate}
\usepackage{mathtools}
\usepackage[colorlinks=true ,urlcolor=blue,urlbordercolor={0 1 1}]{hyperref}

\begin{document}

\title{Entanglement Entropy of Target Functions for Image Classification and Convolutional Neural Network}
\author{%
Ya-Hui Zhang
}
\affiliation{
Department of Physics, Massachusetts Institute of Technology, Cambridge, MA, USA
}

\date{\today}

\begin{abstract}
The success of deep convolutional neural network (CNN) in computer vision especially image classification problems requests a new information theory for function of image, instead of image itself. In this article, after establishing a deep mathematical connection between image classification problem and quantum spin model, we propose to use entanglement entropy, a generalization of classical Boltzmann-Shannon entropy, as a powerful tool to characterize the information needed for representation of general function of image. We prove that there is a sub-volume-law bound for entanglement entropy of  target functions of reasonable image classification problems. Therefore target functions of image classification only occupy a small subspace of the whole Hilbert space.  As a result, a neural network with  polynomial number of parameters is efficient for representation of such  target functions of image. The concept of entanglement entropy can also be useful to characterize the expressive power of different neural networks. For example, we show that to maintain the same expressive power, number of channels $D$ in a convolutional neural network should scale with the number of convolution layers $n_c$ as $D\sim D_0^{\frac{1}{n_c}}$. Therefore, deeper CNN with large $n_c$ is more efficient than shallow ones.
\end{abstract}

\pacs{Valid PACS appear here}
\maketitle

\section{Introduction}
Deep Convolutional Neural Network has achieved great success in computer vision \cite{lecun1995convolutional,lecun1990handwritten,krizhevsky2012imagenet,he2016deep}. However, a complete theoretical understanding of how it works is still absent, despite efforts both numerically \cite{zeiler2014visualizing,yosinski2015understanding} and analytically \cite{bengio2011expressive,eldan2016power,raghu2016expressive}.  Especially, even for simplest black-white image with $N$ pixels, $O(2^N)$ parameters should be necessary to represent a general function of image. However, in practice, convolutional neural network with $O(N^p)$ parameters works quite well in image classification problems. The only way to resolve the above paradox is the following. Target functions of image classification problems occupy only a small subspace of the whole function space and CNN are designed to represent functions in this subspace. Therefore, to further understand which kind of neural network architecture is better, we should first characterize this subspace. There are also other fundamental questions: why does small convolutional kernel work well?  Why is increasing the depth of the CNN more efficient than increasing number of channels at each layer?  In this article, we try to answer these questions using entanglement entropy, one of the most important concepts in modern theoretical physics. 

It's well known that functions of image form a Hilbert Space. However, it's not emphasized before that this Hilbert space has a tensor product structure because of locality of each pixel.  Suppose we have a two dimensional black-white image with size $N=L\times L$.  To preserve locality, we should think of an image as a two dimensional lattice, instead of a  vector with dimension $N$. In lattice representation of image, as we will shown in the main text, the Hilbert Space has a tensor product structure: $\mathcal{H}=\underset{i}{\bigotimes} \mathcal{H}_i$, where $\mathcal{H}_i$ can be thought as a two dimensional local Hilbert space at each pixel. Besides, we will show an amazing mathematical relation: this Hilbert space of functions of image is exactly isomorphic to the Hilbert space of a quantum spin model \cite{wen2004quantum} in the same lattice. Basically up to some normalization factor, any function of image can be thought of as a wavefunction and then has a one to one correspondence with a quantum state of a quantum spin model.  

Quantum spin model has been extensively studied over last several decades and entanglement entropy has been shown to be a powerful tool to characterize a wavefunction in the Hilbert space \cite{eisert2010colloquium}. Despite that the Hilbert Space is exponentially large, tensor network with $O(N^p)$ parameters is efficient to represent a general ground state wavefunction of local Hamiltonian.  The reason is because that the entanglement entropy of these wavefunctions obey an area law bound(with log corrections in some cases).  It turns out that most of functions in the Hilbert space has volume law entanglement entropy and need $O(2^N)$ parameters to represent.  However, locality constrains interesting wavefunctions (wavefunction of ground state) to an exponentially small subspace of the whole Hilbert space. Because of this locality constraint, tensor networks are successful in approximation of these wavefunctions \cite{orus2014practical}.

Because the Hilbert Space of image classification problem is mathematically equivalent to the Hilbert space of quantum spin model, we expect techniques in one field will also have useful applications in another field. Actually Matrix Product State, a special tensor network widely used in quantum physics numerical simulation, has already been shown to also work for MNIST  handwritten-digit recognition classification problem \cite{Stoudenmire_nips2016}. Besides, restricted boltzmann machine developed in computer vision field has been proposed to be a variational ansatz of  the wavefunction of a quantum spin model\cite{carleo2017solving,deng2017quantum,gao2017efficient,huang2017neural}. In this article, we will try to answer a more fundamental question: Can entanglement entropy also be a useful concept for image classification problems and other computer vision problems.

 The Boltzmann-Shannon entropy is a key tool to characterize the information of an image in information theory \cite{shannon2001mathematical} . Now in the new era of artificial intelligence, we need an information theory for function of image, instead of image itself. Entanglement entropy, as a generalization of Boltzmann-Shannon entropy, can be an efficient way to characterize the information needed for representation of a function of image. First, we need to emphasize that the definition of entanglement entropy is not restricted to quantum mechanics. Actually, for any Hilbert Space with local tensor product structure, bipartite entanglement entropy is well defined mathematically. Because functions of image form such a Hilbert Space with tensor product structure, we can always define entanglement entropy. The only question is whether this concept is useful or not. In this article, we will show that entanglement entropy is a useful characterization of difficulty to represent a target function. We will show that entanglement entropy of target functions of image classification problems are bounded by a sub-volume-law(very likely to be area-law for simple problems). Therefore one pixel only entangle locally with pixels nearby. As a result, a neural network with local connection(like convolution kernel) is efficient to represent such a function and only $O(N^p)$ instead of $O(2^N)$ parameters are needed.  Entanglement Entropy can also be a powerful tool to study the expressive power of different network architectures. For example, we will argue that entanglement entropy of a deep convolutional neural network scales as $S\sim n_c\log D$, where $n_c$ is the number of convolution layers and $D$ is the number of hidden channels of each layer.  Therefore, to keep the  entanglement entropy of CNN at the same level as the target function (thus keep the same expression power), number of channels should scale as $D \sim D_0^{\frac{1}{n_c}}$, where $D_0$ is the number of channels for shallow CNN with depth $1$. 

\section{Problem Definition and Hilbert Space \label{section:hilbert}}
In this section we define the image classification problem and discuss the structure of the Hilbert Space of functions of image.

\subsection{Problem Definition of Image Classification}
For simplicity we consider two-class image classification problem. Multi-class classification can be transformed to multiple two-class classification problem. To be specific, we consider the problem of classify whether an image is a cat. Every image has $N=L^2$ pixels and every pixel can be either $0$ or $1$. We define $S$ to be the set of all images and $S$ includes in total $2^N$ images.  We also define the set of all complex value function of image as $\mathcal{H}_I=\{f:S \rightarrow \mathbb{C}\}$.

We assume that there exists a target function of image $F \in H_I$ defined as following:
\begin{equation}
  F(s)=\begin{cases}
  1, & \text{if s is an image of cat.} \\
  0, & \text{otherwise.}
  \end{cases}
  \label{eq:golden}
\end{equation}

For supervised learning, $F(s)$ is known for all training data. Supervised learning is defined as finding a funcition which can approximate target function $F$ well by solving the following optimization problem:
\begin{equation}
  \min_{f \in \mathcal{H}_I} E[f]
  \label{eq:cvopt}
\end{equation}
where $E[f]$ is a functional, a function  on function space , defined as:
\begin{equation}
  E[f]=\sum_{s\in S} V(f(s)-F(s))P(s)
\end{equation}
where $V$ is a cost function and $P(s)$ is a probability distribution of images, which is decided by specific problem. In supervised learning, $P(s)$ can be approximated by dataset $S_{data}$.
\begin{equation}
  E[f]=\sum_{s \in S_{data}} V(f(s)-F(s))
  \label{eq:cost}
\end{equation}

It's hard to solve the optimization problem in Eq.~\ref{eq:cvopt} directly because $E[f]$ is a functional. Thus we should represent $f \in \mathcal{H}_I$ by a specific form with finite number of parameters first.  In computer vision applications, deep convolutional neural network(DCNN) shows good performance to represent $f\in \mathcal{H}_I$:
\begin{equation}
  f|_\omega(s)=DCNN_\omega(s)
\end{equation}
where, $DCNN_\omega$ means a function specified by a deep convolutional neural network with a high dimensional parameter vector $\omega$. In real application, dimension of parameter $\omega$ is a polynomial of $N$: $|\omega| \sim O(N^p)$. 

Then the optimization problem becomes a minimization problem of multi-variable function:
\begin{equation}
  \min_{\omega}E(\omega)=E[f|_\omega]
\end{equation}
which can be solved by gradient descent methods because $DCNN|_\omega$ is differentiable to $\omega$.

\subsection{Hilbert Space $\mathcal{H}_I$ \label{section:basis}}
Next we will show that space of all functions of image $\mathcal{H}_I$ is actually a Hilbert Space with dimension $2^N$. First, it's obvious that this function space is a vector space with definition of addition and scalar multiplication as:
\begin{equation}
  (a f_1 +b f_2)(s)=a f_1(s) + b f_2(s)
\end{equation}
where, $f_1,f_2 \in H_i$ and $a,b \in \mathbb{C}$.

Then we define inner product as:
\begin{equation}
  \left <f_1|f_2 \right >=\sum_{s \in S}f_1^*(s)f_2(s)
\end{equation}
where, $\left<f_1|f_2\right>$ stands for inner product of $f_1,f_2 \in \mathcal{H}_I$. 

It's easy to verify that this definition of inner product satisfies all properties of inner product. Therefore $H_I$ is a Hilbert Space. Next we show that its dimension is $2^N$. First we define $2^N$ functions(vectors) in $H_I$:
\begin{equation}
  e_i(s)=\begin{cases}
  1 & \text{if } s=s_i\\
  0 & \text{otherwise}
  \end{cases}
  \label{eq:basis}
 \end{equation}
 where $i=1,2,...,2^N$ and $s_i \in S$ stands for the $i$th image in $S$. These $\{e_i \in \mathcal{H}_I\}$ are $2^N$ linearly independent vectors in $\mathcal{H}_I$ and it's easy to show that any function $f \in \mathcal{H}_I$ is a linear combination of these $\{e_i\}$:
 \begin{equation}
 f=\sum_{i=1}^{2^N}f(s_i)e_i
 \end{equation}

  Besides, $\left<e_i|e_j\right>=\delta_{ij}$, so $\{e_i\}$ forms an orthogonal basis of $\mathcal{H}_I$. Therefore dimension of $\mathcal{H}_I$ is indeed $2^N$. 

\subsection{Tensor Product Structure of Hilbert Space}
\begin{figure}
\includegraphics[width=150pt]{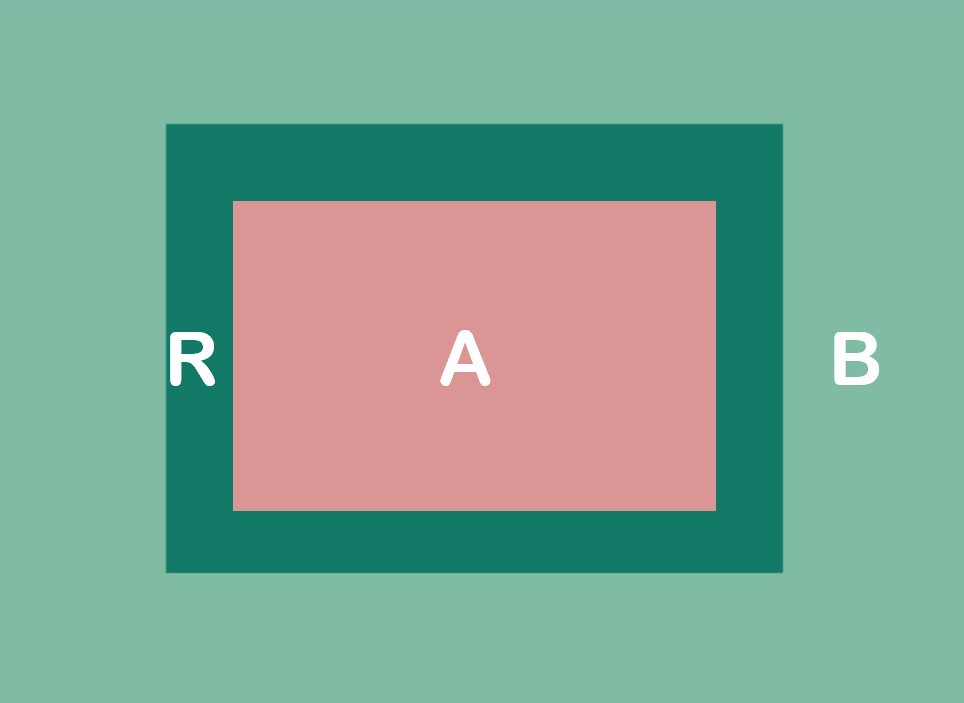}
\caption{Partition of an image. $B$ region is light green. $A$ region includes both dark green and pink part. $R$ region with dark green part is inside $A$.   The function space of the whole image has tensor product structure $\mathcal{H}=\mathcal{H_A}\cong \mathcal{H_B}$. As shown in Section~\ref{section:sub-volume-law}, for target function of reasonable image classification problems, region $B$ entangles with region $A$ only through region $R$ with size $r$.  Therefore the corresponding entanglement entropy satisfies area-law: $S_{AB}\sim r L_{AB}$.}
\label{fig:AB}
\end{figure}

We have shown that functions of size $N$ image form a $2^N$ dimensional Hilbert Space $\mathcal{H}_I$.  Next we will show the tensor product structure of this Hilbert Space. Actually any size $N$ image can be partitioned to two image $A$ and $B$, as shown in Fig.~\ref{fig:AB}. We label the set of images for $A$ and $B$ region as $S_A$ and $S_B$. Then functions of sub-image $S_A$ form a Hilbert Space $\mathcal{H}_A$ with dimension $2^{N_A}$. Similarly, functions of image $B$ form Hilbet Space $\mathcal{H}_B$ with dimension $2^{N_B}$.  Next we will prove that $\mathcal{H}_I=\mathcal{H}_A \otimes \mathcal{H}_B$.

As we show in Section~\ref{section:basis}, $\mathcal{H}_A$ has $2^{N_A}$ basis vectors $e^A_i$, each $i=1,2,...,2^{N_A}$ corresponds to one image $s^A_i\in S_A$, as defined in Eq.~\ref{eq:basis}. Similarly, $\mathcal{H}_B$ has $2^{N_B}$ basis vectors $e^B_j$ with $j=1,2,...,2^{N_B}$ corresponds to $s^B_j \in S_B$. 

For each original image $s\in S$, it can be uniquely decomposed to two sub-images $s^A_i$ and $s^B_j$. We label $s_{ij}\in S$ as $s_{ij}=s^A_i\otimes s^B_j$. Then according to Eq.~\ref{eq:basis}, it corresponds to a basis in $H_I$, $e_{ij}=e^A_i \otimes e^B_j$.  It's easy to verify this definition gives the tensor product structure $\mathcal{H}_I=\mathcal{H}_A \bigotimes \mathcal{H}_B$.

We can further decompose $\mathcal{H}_A$ and $\mathcal{H}_B$ following the above procedure. Finally, we can think of one image $s\in S$ as composed of $N$ images, each of which is just a pixel. As shown above, each pixel $i$ has a two dimensional Hilbert Space $\mathcal{H}_i$ and the total Hilbert Space is a tensor product:
\begin{equation}
\mathcal{H}_I=\underset{i}{\bigotimes} \mathcal{H}_i
\end{equation}

Therefore, the Hilbert Space in the image classification problem has a tensor product structure.  It can also be shown that this Hilbert Space $\mathcal{H}_I$ is mathematically equivalent to the Hilbert Space of  quantum spin model $\mathcal{H}_S$: $\mathcal{H}_I \cong \mathcal{H}_S$. Therefore, each function defined on image set $S$ has a one to one correspondence to a wavefunction of quantum spin model.  Details of this equivalence is shown in Appendix.A. Because of this amazing equivalence, we can use mathematical tools developed in quantum physics field to deal with functions of image in computer vision field.

\section{Entanglement Entropy}
As shown in Section~\ref{section:hilbert}, both computer vision field and quantum physics field is trying to represent a vector in a $2^N$ dimensional Hilbert Space $\mathcal{H}_I\cong \mathcal{H}_S$. Then it's natural to ask the following question: Is it possible to represent a general vector in the $2^N$ dimensional Hilbert Space with $O(N^p)$ number of parameters?  The answer is well known in quantum physics field. It's not possible to approximate a general vector in this Hilbert Space with $O(N^p)$ parameters. We can at best approximate vectors of a subspace whose function has entanglement entropy smaller than volume-law. In practice, ground state wavefunction of local Hamiltonian can be represented by tensor network efficiently. The reason is that entanglement entropy of these functions are bounded by area law, while most of the functions in Hilbert space $\mathcal{H}_S$ have volume law entanglement entropy.

Empirically deep convolutional neural network is successful in image classification problems. We label the set of target functions in image classification problems as $\mathcal{H}_{T}\subset \mathcal{H}_I$. Inspired by quantum physics, we propose that entanglement entropy can also be useful to characterize function in $\mathcal{H}_T$. Especially, we will show that a function $f\in \mathcal{H}_T$ satisfies sub-volume-law bound (very likely to be area-law) for entanglement entropy.

\subsection{Density Matrix}
To make the following analysis easier, we will choose a simple normalization condition. We choose an image $s_0$  with label $1$ as benchmark, and let $|f(s)/f(s_0)|^2$  represent the possibility that $s$ also has label $1$. Then both in image classification and quantum physics, for a function $f$, we only care about ratio $f(s)/f(s')$ for any images $s,s'\in S$. Therefore, for simplicity we can always normalize the norm of the function to be $1$:
\begin{equation}
 \braket{f|f}= \sum_{s\in S}|f(s)|^2=1
 \label{eq:norm}
\end{equation}

Then for every  $f \in \mathcal{H}_I$, we can define a $2^N \times 2^N$ density matrix\cite{nielsen2002quantum} as:
\begin{equation}
  \rho_{ij}=f^*(s_i)f(s_j)
\end{equation}
where $\rho_{ij}$ means the entry $\rho(i,j)$ with $i,j=1,2,...,2^N$. And $s_i$ means the $i$th image in $S$.

It's convenient to write the density matrix in dirac notation as:
\begin{equation}
  \rho=\sum_{ij}f^*(s_i)f(s_j) \ket{s_i}\bra{s_j}
  \label{eq:rhodirac}
\end{equation}
As we have shown in Eq.~\ref{eq:basis}, each basis $e_i$ of the Hilbert Space $\mathcal{H}_I$ uniquely correspond to an image $s_i \in S$. So we use $\ket{s_i}$ to denote this basis vector.  $\bra{s_i}$ is the corresponding covector.  Therefore density matrix in Eq.~\ref{eq:rhodirac} explicitly show that density matrix is  a linear transformation in the Hilbert Space $\mathcal{H}_I$.  Later we will see that properties of density matrix is independent of basis.

Because of normalization condition Eq.~\ref{eq:norm}, we can easily prove that
\begin{equation}
  Tr \rho=1
\end{equation}

 The density matrix $\rho$ can be thought as a generalization of probability distribution with $tr\rho=1$. In computer vision, ratio of diagonal term $\frac{\rho_{ss}}{\rho_{s's'}}=\frac{|f(s)|^2}{|f(s'|^2})$ can be seen as the ratio of probability $\frac{P(s)}{P(s')}$, where $P(s)$ is the probability that $s$ is a cat.

Von Neumann entanglement entropy is defined as a generalization of Boltzmann-Shannon entropy:
\begin{equation}
  S=-tr(\rho \log \rho)
  \label{eq:entropy}
\end{equation}

It can be proven that for any $f \in H_I$,
\begin{equation}
  S=0
\end{equation}

So this definition of entanglement entropy seems to be meaningless. However we will show that entanglement entropy should be defined for a bipartite partition for an image. It's actually a measure of entanglement between two sub-images $A$ and $B$ for a target function $f\in \mathcal{H}_I$. One intuitive understanding is that it characterize nonlinearity of this function $f$ between part $A$ and $B$.

\subsection{Bipartite Entanglement Entropy}
We define bipartite Von Neumann entanglement entropy now. We divide an image to two parts $A$ and $B$, as shown in Fig.~\ref{fig:AB}. Then total Hilbert Space can be decomposed to tensor product state $\mathcal{H}_I=\mathcal{H}_A \otimes \mathcal{H}_B$, where $\mathcal{H}_A$ and $\mathcal{H}_B$ are two Hilbert Space defined on image A and B with dimension $2^{N_A}$ and $2^{N_B}$.

Because any basis $s_i\in S$ correspond to a basis of $\mathcal{H}_I$. We can label it  as $\ket{i}=\ket{i_B}\ket{i_A}$, where $i_A=1,2,...,2^{N_A}$ and $i_B=1,2,...,2^{N_B}$. Here each $i$ can be decomposed to two parts $i_A$ and $i_B$. $\ket{i_A}$ and $\ket{i_B}$ are basis of $\mathcal{H}_A$ and $\mathcal{H}_B$.

As we showed above, the density matrix $\rho$ is basis free. Therefore it's easy to work in the following notation:
\begin{equation}
  \rho=\sum_{i_A,i_B,j_A,j_B}\rho_{i_A,i_B;j_A,j_B} \ket{i_B}\ket{i_A}\bra{j_A}\bra{j_B}
\end{equation}

Then we can get a density matrix defined on $H^A$ by tracing the $H_B$ part:
\begin{align}
   &\rho^A=Tr_B\rho=\sum_{i'_B}\braket{i'_B|\rho|i'_B}\notag\\
   &\ =\sum_{i'_B;i_A,i_B,j_A,j_B}\rho_{i_A,i_B;j_A,j_B}\ket{i_A}\bra{j_A} \braket{i'_B|i_B}\braket{j_B|i'_B}
 \end{align} 

Or equivalently after using $\braket{i_B|i'_B}=\delta_{i_B, i'_B}$, $\rho^A$ is
\begin{equation}
  \rho^A_{i_Aj_A}=\sum_{i_B=1}^{2^{N_B}}\rho_{i_A,i_B;j_A,i_B}
\end{equation}

The density matrix defined on sub image has property:
\begin{equation}
  Tr \rho^A=1
\end{equation}

But it doesn't satisfy $(\rho^{A})^2=\rho^A$ anymore.

More importantly, we can also define entanglement entropy following Eq.~\ref{eq:entropy} as
\begin{equation}
  S_A=-tr(\rho^A \log \rho^A)
\end{equation}

We list several important properties of entanglement entropy.
\begin{property}
For any partition $A$ and $B$, $S_A=S_B=S_{AB}$. 
\end{property}

\begin{property}
If dimension of matrix $\rho^A$ is $D$, $S_A \leq \log D$.
\label{prop:rank}
\end{property}

\begin{property}
$S_{A}$ remains unchanged under unitary transformation $\rho^A \rightarrow U \rho^A U^\dagger$, where $U$ is a unitary matrix.
\end{property}

The last property means that entanglement entropy is basis free. So we can always diagonalize $\rho^A$ because it's hermitian. In the diagonal form, $P_i=\rho^A_{ii}$ forms an ordinary probability distribution and entanglement entropy $S_A$ is just the Shannon entropy.  Then we know $S_{max}=\log D$, where $D$ is the number of nonzero eigenvector of $\rho^A$.

\subsection{Meaning of Entanglement Entropy }
For general function $f\in H_I$, $S_{AB} \neq 0$. To continue, we need to understand the meaning of $S_{AB}$ first.  Let's understand what type of function has zero entanglement entropy as a starting point.

\begin{theorem}
Any function $f \in H_{I}$, if it can be written in a product form: $f=f_A \otimes f_B$ which means that $f(s)=f_A(s_A) \times f_B(s_B)$, where $f_A \in \mathcal{H}_A$ and $f_B \in \mathcal{H}_B$, then $S_{AB}=S_A=S_B=0$.
\label{th:prod}
\end{theorem}

It's easy to prove that $\rho^A_{i_A,j_A}=f_A^{*}(i_A)f_A(j_A)$. Thus $\rho^A$ can be seen as generated from $f_A \in H_A$ and thus $S_A=0$. For this special form of function, part $A$ and $B$ are totally independent and there is no entanglement between these two parts.  

One special case is the famous logistic regression $f(s)\sim e^{\sum_{i=1}^N a_i s(i)}=\prod_{i=1}^N e^{a_i s(i)}$, where $s(i)$ is the value of pixel $i$.  From Theorem~\ref{th:prod}, we know entanglement between any bipartite partition $A$ and $B$ is zero for this function. As a result, the logistic regression is impossible to represent any function of image which has nonzero entanglement between two partitions.

For general function $f\in H_I$, it can be written in Schmidt decomposition form:
\begin{equation}
  f=\sum_{i=1}^m a_i f^i_A \otimes f^i_B
  \label{eq:decomposition}
\end{equation}
where $m\leq \min\{2^{N_A}, 2^{N_B}\}$ and $f^i_A \in H_A$ and $f^i_B \in H_B$. We also have $\sum_{i=1}^m |a_i|^2=1$ Besides, different $f^i_A$ are orthogonal $\braket{f^i_A|f^j_A}=\delta_{ij}$.

After tracing over region $B$, we get
\begin{equation}
  \rho^A=\sum_{i=1}^m |a_i|^2 \ket{f^i_A}\bra{f^i_A}
\end{equation}

Therefore in the basis of $f^i_A$, $\rho^A$ is a diagonal matrix with diagonal element $|a_i|^2$. We have the following theorem:

\begin{theorem}
For any function $f\in \mathcal{H}_I$ and any partition of $A$ and $B$, entanglement entropy is bounded by volume law $S_{AB}\leq \min\{N_A,N_B\} \log 2$.
\label{th:prod}
\end{theorem}

The theorem comes naturally from Property~\ref{prop:rank} of Entanglement Entropy. We have the following definition.

\begin{definition}
For a function $f\in \mathcal{H}_I$, it has volume-law entanglement entropy if $S_{AB} \sim O(L_{AB})$ for any bipartite partition $A$ and $B$ , where $L_{AB}$ is the length of boundary between $A$ and $B$.
\label{def:volume-law}
\end{definition}

 Volume law can be intuitively understood as following. For most of functions $f \in \mathcal{H}_I$, two partitions $A$ and $B$ are not independent and entangled. As a result, $\rho^A$ can not be generated by one single function, and is dependent on state of pixels of region $B$. In general, part $B$ has $2^{N_B}$ possible states and we need $O(2^{N_B})$ independent functions to describe $\rho^A$.  Thus in the diagonal form of $\rho^A$, there are $O(2^{N_B})$ nonzero diagonal elements and thus $S_{max}\sim \log (2^{N_B})\sim N_B$.  

One intuitive understanding of entanglement entropy is the range of pixels entangled with one pixel. In the case of volume law, one pixel in part $A$ is entangled with every pixel in part $B$ and thus total entanglement entropy is proportional to the number of pixels, and thus a volume law.  To represent a function of volume law, fully connected network is necessary and local connection like convolutional kernel is apparently impossible to represent such a function.

$O(2^N)$ number of parameters is necessary to represent volume-law function. However, for function with area-law entanglement entropy, $O(N^p)$ number of parameters may be enough to represent it.

\begin{definition}
For a function $f\in \mathcal{H}_I$, it has area-law entanglement entropy if $S_{AB} \sim O(L_{AB})$ for any bipartite partition $A$ and $B$ , where $L_{AB}$ is the length of boundary between $A$ and $B$.
\label{def:area-law}
\end{definition}

Area-law entanglement entropy implies that one pixel is only locally entangled with pixels in its neighborhood. Thus entanglement for part $A$ and $B$ is only from boundary and thus is proportional to $O(L_{AB})$. Therefore, to represent a function with area-law entanglement entropy, we only need local connections between pixels, such as convolutional kernel with small width.

\subsection{Examples of Volume-Law and Area-Law Image Classification Problem}
In this section we provide two examples of image classification. We will show that target function of one problem has volume-law entanglement entropy and the other one has area-law entanglement entropy. 
\subsubsection{Volume-Law Example: Random Image Set}
Considering the following image classification problem. We randomly generate a set  of images $S_I$ and label these images as $1$. Other images not in this set are labeled as $0$. Then the image classification problem is to supervised-learning this set $S_I$. One can imagine that this task is impossible for any neural network because these images in set $S_I$ don't have any pattern at all.

Next we give a quantitative statement of no-pattern by showing that the corresponding target function has volume-law entanglement entropy. The target function as defined in Eq.~\ref{eq:golden} can be thought as a random  vector in the Hilbert Space $\mathcal{H}_{I}$. It has been shown that a random chosen vector in the Hilbert Space $\mathcal{H}_I \cong \mathcal{H}_S$ has volume-law entanglement entropy because it correspond to a thermalized state with almost infinite temperature \cite{page1993average}. Because the target function has volume law entanglement entropy, it's impossible to represent with a simple locally connected neural network. To represent such a function, long range connection with exponentially large number of parameters is necessary.

\subsubsection{Area-Law Example: Recognizing Closed Loops}
We give an example of image classification problem with area-law entangled target function. The task is closed loop recognition. If an image only has closed loops, the label is $1$. If there is any open string in the image, the label is $0$. This task could be efficiently accomplished by training simple convolutional neural network. The target function of this problem can be analytically proven to have area-law entanglement entropy $S_{AB}\sim L_{AB} \log 2$ because it correspond to the famous quantum loop gas state in toric code \cite{kitaev2003fault}. The intuition is that to decide whether a line is a closed loop or open string, it's only necessary to check some local constraints. As a result, pixels entangle locally and a small convolutional kernel can be used for this problem.

\section{Sub-Volume-Law Entanglement Entropy of Target Functions for Image Classification \label{section:sub-volume-law}}
We have already seen that entanglement entropy is a powerful tool to measure the difficulty for representing a function. Functions with volume law entanglement entropy generally need $O(2^N)$ parameters to approximate and functions with area-law entanglement entropy are possible to be approximated by short-range connected networks with $O(N^p)$ parameters.  As CNN is quite successful in image classifications, it's natural to conjecture that objective functions in image classification problems are area-law entangled.  Next we will justify this conjecture.

Suppose we have a image classification problem. The target function is that $F(s)=1$ if $s$ has label 1 and $F(s)=0$ otherwise. We have a partition $A$ and $B$. The boundary is included in region $A$ and the length of boundary is $L_{AB}$. We label the set of images with label $1$ as $S_{I}$.

Then density matrix is:
\begin{equation}
  \rho=\frac{1}{N_I}\sum_{s,s' \in S_{I}}\ket{s_A}\bra{s'_A} \otimes \ket{s_B}\bra{s'_B}
\end{equation}
where  $N_I$ is the number of label $1$ images. $s_A,s'_A \in S_A$  and $s_B,s'_B \in S_B$. $S_{A}$ ($S_B$) are set of images in region $A$ ($B$).

Next we trace over $H_B$:
\begin{align}
  \rho^A&=tr_B \rho\notag\\
  &=\sum_{s^{''}_B}\bra{s^{''}_B}\rho\ket{s^{''}_B}\notag\\
  &=\frac{1}{N}\sum_{s,s' \in S_{cat}}\ket{s_A}\bra{s'_A} \sum_{s^{''}_B}\braket{s^{''}_B | s_B}\braket{s'_B|s^{''}_B} \notag\\
  &=\frac{1}{N}\sum_{s,s' \in S_{cat}}\ket{s_A}\bra{s'_A} \sum_{s^{''}_B}\delta_{s^{''}_Bs_B}\delta_{s^{''}_Bs'_B} \notag\\
  &=\sum_{s_A \sim s'_A}\frac{N_{s_As'_A}}{N} \ket{s^A}\bra{s'_A}
\end{align}
where $s_A \sim s'_A$ means that there exits $N_{s_As'_A}\geq 1$ possible $s_B \in S_B$  which  can generate label $1$ image by combining with both $s_A$ and $s'_A$.

Naively $\rho^A$ is a $2^{N_A} \times 2^{N_A}$ matrix. However, we can organize it to $2^{L_{AB}}$ blocks with the following natural assumption.

\begin{assumption}
If two label $1$ images are exactly the same in region $B$, they must also be the same at the boundary.
\end{assumption}

The assumption follows naturally from the continuation of part $B$ if the image classification problem is for an object which is smooth locally.  With this assumption, we know that $s_A \sim s'_A$ only if they have the same boundary.  There are $2^{L_{AB}}$ possible states of the boundary. Therefore the density matrix $\rho^A$ can be organized to $2^{L_{AB}}$ blocks, each of which correspond to one state of the boundary.b

However, $s_A\sim s'_A$ may not hold even if $s_A$ and $s'_A$ have the same boundary.  We need another assumption of the image classification. We label a subregion with range $r$ within region $A$ close to the boundary as $R$, as shown in Fig~\ref{fig:AB}. In the following, we will assume that whether $s_A \sim s'_A$ and $N_{s_A s'_A}$ only depends on the state at this region $R$. We label $S_{s_R}$ as the set of $s_A \in S_A$ which is the same as $s_R$ in region $R$ and can be extended to a label $1$ image by some $s_B \in S_B$.

\begin{assumption}
If $s_A\in S_{s_R}$ and $s'_A \in S_{s'_R}$, then $N_{s_A s'_A}=N_{s_R s'_R}$ which only depends on their state at region $R$.
\end{assumption}

The above assumption apparently holds for $r=L_A$. However, we expect $r=O(1)$ for simple image classification problems with locality. The assumption means that whether two images can extend to a label $1$ image with the same $B$ part only depends on their states on region $R$ and doesn't depend on inner region $A/R$. The assumption is true because the image in $B$ part is smooth extension of $R$ region.  With this assumption, We define the following basis:
\begin{equation}
  \ket{\Psi_{s_R}}=A \sum_{s_A \in S_{s_R}} \ket{s_A}
\end{equation}
where $A$ is a normalization factor.

There are $2^{r L_{AB}}$ possible $s_R$. In terms of these $2^{r L_{AB}}$  orthogonal basis
\begin{equation}
  \rho^A=\sum_{s_R s'_R}\frac{N_{s_R s'_R}}{N_I}\ket{\Psi_{s_R}}\bra{\Psi_{s'_R}}
  \label{eq:prerhoa}
\end{equation}

$\rho^A$ is a matrix with dimension $2^{r L_{AB}}$. Then we know immediately that entanglement entropy $S_A\leq r L_{AB} \log 2$.

\begin{theorem}
For image classification problem satisfying the above two assumptions, the entanglement entropy for target function is bounded by $S_{AB} \leq r L_{AB} \log 2$. $r$ is a characterization of the range of entanglement of each image classification problem.
\end{theorem}

Thus $r$ can be thought of as the range of entanglement. For such a target function, a pixel only entangles with pixels within the distance of $r$. Note that, entanglement between pixels are defined for a function $f\in \mathcal{H}_I$. We are meaning that, to approximate such a function $f$, we need one pixel to entangle with other pixels in the network (whatever the network is, convolutional neural network, tensor network, or a network not proposed yet).

In summary, we argued that the entanglement entropy of target functions of image classification problems are bounded by a sub-volume-law $S_{AB} \sim r L_{AB}$. $r$ can be seen as a characterization of the difficulty of each classification problems. For simple task like MNIST(hand-written  digit recognition) data, $r \sim O(1)$ is a reasonable estimation and the target function should have area-law entanglement entropy. Some complicated tasks may have $r\sim L^\alpha$ with $\alpha<1$.  But we believe volume law entanglement entropy with $r\sim L$ is very rare because of locality.

It's hard to  analytically extract $r$ for each image classification problem. But numerical calculation of entanglement entropy for each image classification problem may be possible. We leave it to future work.

\section{Application to Convolutional Neural Network}
After showing that entanglement entropy of target function of a image classification problem is bounded by a sub-volume-law $S_{AB}\sim r L_{AB}$. Next we will use entanglement entropy to characterize the expressive power of different neural network architectures.

Specifically we consider deep CNN with $n_p$ pooling layers and $n_c$ convolution layers between pooling layers. The number of channels at each layer is denoted as $D$. For simplicity we assume each layer has the same $D$.  The architecture of CNN is very similar to Multiscale Entanglement Renormalization Ansatz(MERA) \cite{vidal2007entanglement}.  It's reasonable that convolutional neural network is also doing entanglement renormalization as MERA.  Pooling layer of CNN is similar to a block-spin renormalization group step.  For image classification problem with scale invariance, we need $n_p \sim \log L$ to get a scale invariant ansatz with correlation length $\xi \sim L$. 

In practice it's also found that CNN with larger $n_c$ works better while the size of convolution kernel $W$ can be small.  We can understand the role of convolutional layer as a disentangler in MERA. Before pooling layer which reduce the size of a block to $1$, we must extract most important features for this  block and its neighbors to reduce the information lost during pooling layer(RG) process. In a formal language, we must keep the entanglement between each block and its neighbors.  Because entanglement entropy of the target function is smaller than volume law(very likely to be area-law), pixels are only locally entangled. Thus a fully connected network is not necessary and we can use a small convolution kernel.  The number of channels $D$ is similar to the bond dimension in MERA. As shown by numerical experiments, each channel represents a feature of original kernel at previous layer \cite{zeiler2014visualizing}. In quantum physics language, each channel represents a disentangled state of the corresponding kernel, which is exactly the role played by disentangler in MERA. The CNN will be trained to extract $D$ most important features of this kernel. In a quantitative language, it's trained to change the original basis of this kernel to $D$ new vectors, which minimizes the loss of entanglement that will be lost during the following pooling process.  By making analog with MERA, the entanglement entropy of a CNN scales as $\frac{S_{AB}}{L_{AB}} \sim n_c \log D$ \cite{vidal2008class}.  We want the entanglement entropy of the CNN to be at the same level of the target function.  Then $n_c \log D \sim r$ is needed to represent a target function with entanglement entropy $S_{AB} \sim r L_{AB}$.  In another word, we need $D \sim D_0^{\frac{1}{n_c}}$ to keep the expression power of the CNN.  It's then obvious that increasing $n_c$ is much more efficient than increasing $D$.

\section{conclusion}
In conclusion, we propose to use entanglement entropy to characterize the information needed to represent a target function for an image classification problem. We show that entanglement entropy is bounded by sub-volume law (even area-law) for target functions in image classification problems because of locality.  Therefore $O(N^p)$ parameters are enough to represent a target function.  We can also use entanglement entropy to characterize the expressive power of a neural network architecture. Specifically, we show the entanglement entropy of a deep CNN scales as $S_{AB}/L_{AB} \sim n_c \log D$. Therefore a deeper CNN with larger $n_c$ is much more efficient than shallow ones.

A lot of directions are open for future work.  First, numerical techniques should be developed to measure entanglement entropy for each image classification problem and other computer vision problems. Second, as we have shown a deep connection between quantum physics and image classification, ideas and methods in one field may have applications in the other field. Finally, this article is focused on problem of image, which has a spatial lattice.  Time series data is involved in speech recognition and natural language processing problem.  It remains an open question whether we can also characterize functions of time series using entanglement entropy or similar concepts. 
\section{Acknowledgement}
We would like to thank T.Senthil, Roger G.Melko and Yijia Zhang for preview and helpful comments on the manuscript. We also thank Liujun Zou, Michael Pretko, Zhehao Dai,Yang Qi, Li Jing, Zheng Ma, Liyang Xiong, Kang Yang for useful discussions. Especially thank Yan Liu for help on making the plot. This research was supported  by the Simons Foundation through a Simons Investigator Award to Senthil Todadri.
\bibliographystyle{apsrev4-1}
\bibliography{ref}

\begin{thebibliography}{23}%
\makeatletter
\providecommand \@ifxundefined [1]{%
 \@ifx{#1\undefined}
}%
\providecommand \@ifnum [1]{%
 \ifnum #1\expandafter \@firstoftwo
 \else \expandafter \@secondoftwo
 \fi
}%
\providecommand \@ifx [1]{%
 \ifx #1\expandafter \@firstoftwo
 \else \expandafter \@secondoftwo
 \fi
}%
\providecommand \natexlab [1]{#1}%
\providecommand \enquote  [1]{``#1''}%
\providecommand \bibnamefont  [1]{#1}%
\providecommand \bibfnamefont [1]{#1}%
\providecommand \citenamefont [1]{#1}%
\providecommand \href@noop [0]{\@secondoftwo}%
\providecommand \href [0]{\begingroup \@sanitize@url \@href}%
\providecommand \@href[1]{\@@startlink{#1}\@@href}%
\providecommand \@@href[1]{\endgroup#1\@@endlink}%
\providecommand \@sanitize@url [0]{\catcode `\\12\catcode `\$12\catcode
  `\&12\catcode `\#12\catcode `\^12\catcode `\_12\catcode `\%12\relax}%
\providecommand \@@startlink[1]{}%
\providecommand \@@endlink[0]{}%
\providecommand \url  [0]{\begingroup\@sanitize@url \@url }%
\providecommand \@url [1]{\endgroup\@href {#1}{\urlprefix }}%
\providecommand \urlprefix  [0]{URL }%
\providecommand \Eprint [0]{\href }%
\providecommand \doibase [0]{http://dx.doi.org/}%
\providecommand \selectlanguage [0]{\@gobble}%
\providecommand \bibinfo  [0]{\@secondoftwo}%
\providecommand \bibfield  [0]{\@secondoftwo}%
\providecommand \translation [1]{[#1]}%
\providecommand \BibitemOpen [0]{}%
\providecommand \bibitemStop [0]{}%
\providecommand \bibitemNoStop [0]{.\EOS\space}%
\providecommand \EOS [0]{\spacefactor3000\relax}%
\providecommand \BibitemShut  [1]{\csname bibitem#1\endcsname}%
\let\auto@bib@innerbib\@empty
\bibitem [{\citenamefont {LeCun}\ \emph {et~al.}(1995)\citenamefont {LeCun},
  \citenamefont {Bengio} \emph {et~al.}}]{lecun1995convolutional}%
  \BibitemOpen
  \bibfield  {author} {\bibinfo {author} {\bibfnamefont {Y.}~\bibnamefont
  {LeCun}}, \bibinfo {author} {\bibfnamefont {Y.}~\bibnamefont {Bengio}},
  \emph {et~al.},\ }\href@noop {} {\bibfield  {journal} {\bibinfo  {journal}
  {The handbook of brain theory and neural networks}\ }\textbf {\bibinfo
  {volume} {3361}},\ \bibinfo {pages} {1995} (\bibinfo {year}
  {1995})}\BibitemShut {NoStop}%
\bibitem [{\citenamefont {LeCun}\ \emph {et~al.}(1990)\citenamefont {LeCun},
  \citenamefont {Boser}, \citenamefont {Denker}, \citenamefont {Henderson},
  \citenamefont {Howard}, \citenamefont {Hubbard},\ and\ \citenamefont
  {Jackel}}]{lecun1990handwritten}%
  \BibitemOpen
  \bibfield  {author} {\bibinfo {author} {\bibfnamefont {Y.}~\bibnamefont
  {LeCun}}, \bibinfo {author} {\bibfnamefont {B.~E.}\ \bibnamefont {Boser}},
  \bibinfo {author} {\bibfnamefont {J.~S.}\ \bibnamefont {Denker}}, \bibinfo
  {author} {\bibfnamefont {D.}~\bibnamefont {Henderson}}, \bibinfo {author}
  {\bibfnamefont {R.~E.}\ \bibnamefont {Howard}}, \bibinfo {author}
  {\bibfnamefont {W.~E.}\ \bibnamefont {Hubbard}}, \ and\ \bibinfo {author}
  {\bibfnamefont {L.~D.}\ \bibnamefont {Jackel}},\ }in\ \href@noop {} {\emph
  {\bibinfo {booktitle} {Advances in neural information processing systems}}}\
  (\bibinfo {year} {1990})\ pp.\ \bibinfo {pages} {396--404}\BibitemShut
  {NoStop}%
\bibitem [{\citenamefont {Krizhevsky}\ \emph {et~al.}(2012)\citenamefont
  {Krizhevsky}, \citenamefont {Sutskever},\ and\ \citenamefont
  {Hinton}}]{krizhevsky2012imagenet}%
  \BibitemOpen
  \bibfield  {author} {\bibinfo {author} {\bibfnamefont {A.}~\bibnamefont
  {Krizhevsky}}, \bibinfo {author} {\bibfnamefont {I.}~\bibnamefont
  {Sutskever}}, \ and\ \bibinfo {author} {\bibfnamefont {G.~E.}\ \bibnamefont
  {Hinton}},\ }in\ \href@noop {} {\emph {\bibinfo {booktitle} {Advances in
  neural information processing systems}}}\ (\bibinfo {year} {2012})\ pp.\
  \bibinfo {pages} {1097--1105}\BibitemShut {NoStop}%
\bibitem [{\citenamefont {He}\ \emph {et~al.}(2016)\citenamefont {He},
  \citenamefont {Zhang}, \citenamefont {Ren},\ and\ \citenamefont
  {Sun}}]{he2016deep}%
  \BibitemOpen
  \bibfield  {author} {\bibinfo {author} {\bibfnamefont {K.}~\bibnamefont
  {He}}, \bibinfo {author} {\bibfnamefont {X.}~\bibnamefont {Zhang}}, \bibinfo
  {author} {\bibfnamefont {S.}~\bibnamefont {Ren}}, \ and\ \bibinfo {author}
  {\bibfnamefont {J.}~\bibnamefont {Sun}},\ }in\ \href@noop {} {\emph {\bibinfo
  {booktitle} {Proceedings of the IEEE conference on computer vision and
  pattern recognition}}}\ (\bibinfo {year} {2016})\ pp.\ \bibinfo {pages}
  {770--778}\BibitemShut {NoStop}%
\bibitem [{\citenamefont {Zeiler}\ and\ \citenamefont
  {Fergus}(2014)}]{zeiler2014visualizing}%
  \BibitemOpen
  \bibfield  {author} {\bibinfo {author} {\bibfnamefont {M.~D.}\ \bibnamefont
  {Zeiler}}\ and\ \bibinfo {author} {\bibfnamefont {R.}~\bibnamefont
  {Fergus}},\ }in\ \href@noop {} {\emph {\bibinfo {booktitle} {European
  conference on computer vision}}}\ (\bibinfo {organization} {Springer},\
  \bibinfo {year} {2014})\ pp.\ \bibinfo {pages} {818--833}\BibitemShut
  {NoStop}%
\bibitem [{\citenamefont {Yosinski}\ \emph {et~al.}(2015)\citenamefont
  {Yosinski}, \citenamefont {Clune}, \citenamefont {Nguyen}, \citenamefont
  {Fuchs},\ and\ \citenamefont {Lipson}}]{yosinski2015understanding}%
  \BibitemOpen
  \bibfield  {author} {\bibinfo {author} {\bibfnamefont {J.}~\bibnamefont
  {Yosinski}}, \bibinfo {author} {\bibfnamefont {J.}~\bibnamefont {Clune}},
  \bibinfo {author} {\bibfnamefont {A.}~\bibnamefont {Nguyen}}, \bibinfo
  {author} {\bibfnamefont {T.}~\bibnamefont {Fuchs}}, \ and\ \bibinfo {author}
  {\bibfnamefont {H.}~\bibnamefont {Lipson}},\ }\href@noop {} {\bibfield
  {journal} {\bibinfo  {journal} {arXiv preprint arXiv:1506.06579}\ } (\bibinfo
  {year} {2015})}\BibitemShut {NoStop}%
\bibitem [{\citenamefont {Bengio}\ and\ \citenamefont
  {Delalleau}(2011)}]{bengio2011expressive}%
  \BibitemOpen
  \bibfield  {author} {\bibinfo {author} {\bibfnamefont {Y.}~\bibnamefont
  {Bengio}}\ and\ \bibinfo {author} {\bibfnamefont {O.}~\bibnamefont
  {Delalleau}},\ }in\ \href@noop {} {\emph {\bibinfo {booktitle} {Algorithmic
  Learning Theory}}}\ (\bibinfo {organization} {Springer},\ \bibinfo {year}
  {2011})\ pp.\ \bibinfo {pages} {18--36}\BibitemShut {NoStop}%
\bibitem [{\citenamefont {Eldan}\ and\ \citenamefont
  {Shamir}(2016)}]{eldan2016power}%
  \BibitemOpen
  \bibfield  {author} {\bibinfo {author} {\bibfnamefont {R.}~\bibnamefont
  {Eldan}}\ and\ \bibinfo {author} {\bibfnamefont {O.}~\bibnamefont {Shamir}},\
  }in\ \href@noop {} {\emph {\bibinfo {booktitle} {Conference on Learning
  Theory}}}\ (\bibinfo {year} {2016})\ pp.\ \bibinfo {pages}
  {907--940}\BibitemShut {NoStop}%
\bibitem [{\citenamefont {Raghu}\ \emph {et~al.}(2016)\citenamefont {Raghu},
  \citenamefont {Poole}, \citenamefont {Kleinberg}, \citenamefont {Ganguli},\
  and\ \citenamefont {Sohl-Dickstein}}]{raghu2016expressive}%
  \BibitemOpen
  \bibfield  {author} {\bibinfo {author} {\bibfnamefont {M.}~\bibnamefont
  {Raghu}}, \bibinfo {author} {\bibfnamefont {B.}~\bibnamefont {Poole}},
  \bibinfo {author} {\bibfnamefont {J.}~\bibnamefont {Kleinberg}}, \bibinfo
  {author} {\bibfnamefont {S.}~\bibnamefont {Ganguli}}, \ and\ \bibinfo
  {author} {\bibfnamefont {J.}~\bibnamefont {Sohl-Dickstein}},\ }\href@noop {}
  {\bibfield  {journal} {\bibinfo  {journal} {arXiv preprint arXiv:1606.05336}\
  } (\bibinfo {year} {2016})}\BibitemShut {NoStop}%
\bibitem [{\citenamefont {Wen}(2004)}]{wen2004quantum}%
  \BibitemOpen
  \bibfield  {author} {\bibinfo {author} {\bibfnamefont {X.-G.}\ \bibnamefont
  {Wen}},\ }\href@noop {} {\emph {\bibinfo {title} {Quantum field theory of
  many-body systems: from the origin of sound to an origin of light and
  electrons}}}\ (\bibinfo  {publisher} {Oxford University Press on Demand},\
  \bibinfo {year} {2004})\BibitemShut {NoStop}%
\bibitem [{\citenamefont {Eisert}\ \emph {et~al.}(2010)\citenamefont {Eisert},
  \citenamefont {Cramer},\ and\ \citenamefont {Plenio}}]{eisert2010colloquium}%
  \BibitemOpen
  \bibfield  {author} {\bibinfo {author} {\bibfnamefont {J.}~\bibnamefont
  {Eisert}}, \bibinfo {author} {\bibfnamefont {M.}~\bibnamefont {Cramer}}, \
  and\ \bibinfo {author} {\bibfnamefont {M.~B.}\ \bibnamefont {Plenio}},\
  }\href@noop {} {\bibfield  {journal} {\bibinfo  {journal} {Reviews of Modern
  Physics}\ }\textbf {\bibinfo {volume} {82}},\ \bibinfo {pages} {277}
  (\bibinfo {year} {2010})}\BibitemShut {NoStop}%
\bibitem [{\citenamefont {Or{\'u}s}(2014)}]{orus2014practical}%
  \BibitemOpen
  \bibfield  {author} {\bibinfo {author} {\bibfnamefont {R.}~\bibnamefont
  {Or{\'u}s}},\ }\href@noop {} {\bibfield  {journal} {\bibinfo  {journal}
  {Annals of Physics}\ }\textbf {\bibinfo {volume} {349}},\ \bibinfo {pages}
  {117} (\bibinfo {year} {2014})}\BibitemShut {NoStop}%
\bibitem [{\citenamefont {Stoudenmire}\ and\ \citenamefont
  {Schwab}(2016)}]{Stoudenmire_nips2016}%
  \BibitemOpen
  \bibfield  {author} {\bibinfo {author} {\bibfnamefont {E.}~\bibnamefont
  {Stoudenmire}}\ and\ \bibinfo {author} {\bibfnamefont {D.~J.}\ \bibnamefont
  {Schwab}},\ }in\ \href
  {http://papers.nips.cc/paper/6211-supervised-learning-with-tensor-networks.pdf}
  {\emph {\bibinfo {booktitle} {Advances in Neural Information Processing
  Systems 29}}},\ \bibinfo {editor} {edited by\ \bibinfo {editor}
  {\bibfnamefont {D.~D.}\ \bibnamefont {Lee}}, \bibinfo {editor} {\bibfnamefont
  {M.}~\bibnamefont {Sugiyama}}, \bibinfo {editor} {\bibfnamefont {U.~V.}\
  \bibnamefont {Luxburg}}, \bibinfo {editor} {\bibfnamefont {I.}~\bibnamefont
  {Guyon}}, \ and\ \bibinfo {editor} {\bibfnamefont {R.}~\bibnamefont
  {Garnett}}}\ (\bibinfo  {publisher} {Curran Associates, Inc.},\ \bibinfo
  {year} {2016})\ pp.\ \bibinfo {pages} {4799--4807}\BibitemShut {NoStop}%
\bibitem [{\citenamefont {Carleo}\ and\ \citenamefont
  {Troyer}(2017)}]{carleo2017solving}%
  \BibitemOpen
  \bibfield  {author} {\bibinfo {author} {\bibfnamefont {G.}~\bibnamefont
  {Carleo}}\ and\ \bibinfo {author} {\bibfnamefont {M.}~\bibnamefont
  {Troyer}},\ }\href@noop {} {\bibfield  {journal} {\bibinfo  {journal}
  {Science}\ }\textbf {\bibinfo {volume} {355}},\ \bibinfo {pages} {602}
  (\bibinfo {year} {2017})}\BibitemShut {NoStop}%
\bibitem [{\citenamefont {Deng}\ \emph {et~al.}(2017)\citenamefont {Deng},
  \citenamefont {Li},\ and\ \citenamefont {Sarma}}]{deng2017quantum}%
  \BibitemOpen
  \bibfield  {author} {\bibinfo {author} {\bibfnamefont {D.-L.}\ \bibnamefont
  {Deng}}, \bibinfo {author} {\bibfnamefont {X.}~\bibnamefont {Li}}, \ and\
  \bibinfo {author} {\bibfnamefont {S.~D.}\ \bibnamefont {Sarma}},\ }\href@noop
  {} {\bibfield  {journal} {\bibinfo  {journal} {Physical Review X}\ }\textbf
  {\bibinfo {volume} {7}},\ \bibinfo {pages} {021021} (\bibinfo {year}
  {2017})}\BibitemShut {NoStop}%
\bibitem [{\citenamefont {Gao}\ and\ \citenamefont
  {Duan}(2017)}]{gao2017efficient}%
  \BibitemOpen
  \bibfield  {author} {\bibinfo {author} {\bibfnamefont {X.}~\bibnamefont
  {Gao}}\ and\ \bibinfo {author} {\bibfnamefont {L.-M.}\ \bibnamefont {Duan}},\
  }\href@noop {} {\bibfield  {journal} {\bibinfo  {journal} {arXiv preprint
  arXiv:1701.05039}\ } (\bibinfo {year} {2017})}\BibitemShut {NoStop}%
\bibitem [{\citenamefont {Huang}\ and\ \citenamefont
  {Moore}(2017)}]{huang2017neural}%
  \BibitemOpen
  \bibfield  {author} {\bibinfo {author} {\bibfnamefont {Y.}~\bibnamefont
  {Huang}}\ and\ \bibinfo {author} {\bibfnamefont {J.~E.}\ \bibnamefont
  {Moore}},\ }\href@noop {} {\bibfield  {journal} {\bibinfo  {journal} {arXiv
  preprint arXiv:1701.06246}\ } (\bibinfo {year} {2017})}\BibitemShut {NoStop}%
\bibitem [{\citenamefont {Shannon}(2001)}]{shannon2001mathematical}%
  \BibitemOpen
  \bibfield  {author} {\bibinfo {author} {\bibfnamefont {C.~E.}\ \bibnamefont
  {Shannon}},\ }\href@noop {} {\bibfield  {journal} {\bibinfo  {journal} {ACM
  SIGMOBILE Mobile Computing and Communications Review}\ }\textbf {\bibinfo
  {volume} {5}},\ \bibinfo {pages} {3} (\bibinfo {year} {2001})}\BibitemShut
  {NoStop}%
\bibitem [{\citenamefont {Nielsen}\ and\ \citenamefont
  {Chuang}(2002)}]{nielsen2002quantum}%
  \BibitemOpen
  \bibfield  {author} {\bibinfo {author} {\bibfnamefont {M.~A.}\ \bibnamefont
  {Nielsen}}\ and\ \bibinfo {author} {\bibfnamefont {I.}~\bibnamefont
  {Chuang}},\ }\href@noop {} {\enquote {\bibinfo {title} {Quantum computation
  and quantum information},}\ } (\bibinfo {year} {2002})\BibitemShut {NoStop}%
\bibitem [{\citenamefont {Page}(1993)}]{page1993average}%
  \BibitemOpen
  \bibfield  {author} {\bibinfo {author} {\bibfnamefont {D.~N.}\ \bibnamefont
  {Page}},\ }\href@noop {} {\bibfield  {journal} {\bibinfo  {journal} {Physical
  review letters}\ }\textbf {\bibinfo {volume} {71}},\ \bibinfo {pages} {1291}
  (\bibinfo {year} {1993})}\BibitemShut {NoStop}%
\bibitem [{\citenamefont {Kitaev}(2003)}]{kitaev2003fault}%
  \BibitemOpen
  \bibfield  {author} {\bibinfo {author} {\bibfnamefont {A.~Y.}\ \bibnamefont
  {Kitaev}},\ }\href@noop {} {\bibfield  {journal} {\bibinfo  {journal} {Annals
  of Physics}\ }\textbf {\bibinfo {volume} {303}},\ \bibinfo {pages} {2}
  (\bibinfo {year} {2003})}\BibitemShut {NoStop}%
\bibitem [{\citenamefont {Vidal}(2007)}]{vidal2007entanglement}%
  \BibitemOpen
  \bibfield  {author} {\bibinfo {author} {\bibfnamefont {G.}~\bibnamefont
  {Vidal}},\ }\href@noop {} {\bibfield  {journal} {\bibinfo  {journal}
  {Physical review letters}\ }\textbf {\bibinfo {volume} {99}},\ \bibinfo
  {pages} {220405} (\bibinfo {year} {2007})}\BibitemShut {NoStop}%
\bibitem [{\citenamefont {Vidal}(2008)}]{vidal2008class}%
  \BibitemOpen
  \bibfield  {author} {\bibinfo {author} {\bibfnamefont {G.}~\bibnamefont
  {Vidal}},\ }\href@noop {} {\bibfield  {journal} {\bibinfo  {journal}
  {Physical review letters}\ }\textbf {\bibinfo {volume} {101}},\ \bibinfo
  {pages} {110501} (\bibinfo {year} {2008})}\BibitemShut {NoStop}%
\end{thebibliography}%

\onecolumngrid
\appendix

\section{Hilbert Space of Quantum Spin Model}

\subsection{Equivalence between $\mathcal{H}_I$ and $\mathcal{H}_S$}
Computer vision is dealing with a Hilbert Space $\mathcal{H}_I$ with dimension $2^N$. In quantum physics, quantum spin model is also on a $2^N$ dimensional Hilbert Space $\mathcal{H}_S$. The basis $\ket{s}$ of $\mathcal{H}_S$ can be thought of as a image $s \in S$. In this section we will further show that $\mathcal{H}_I$ and $\mathcal{H}_S$ are equivalent. In a more precise mathematical language,
\begin{equation}
  \mathcal{H}_I \cong \mathcal{H}_S
\end{equation}

The isomorphism is mathematically easy to prove because two vector space with the same finite number of dimension is isomorphic. $\mathcal{H}_I$ has $2^N$ orthogonal basis $\{e_i\}$ and $\mathcal{H}_S$ has $2^N$ orthogonal basis $\{\ket{s_i}|s_i\in S\}$. We can define a linear transformation $T: \mathcal{H}_I \rightarrow \mathcal{H}_S$ as:
\begin{equation}
  T(e_i)=\ket{s_i}
\end{equation}

Under this definition, a vector $f \in \mathcal{H}_I$ transforms to a $\ket{\Psi} \in \mathcal{H}_S$ under $T$:
\begin{equation}
  \ket{\Psi}=\sum_{s\in S}f(s)\ket{s}
\end{equation}

We can also show that inner product doesn't change under $T$:
\begin{align}
  \left<\Psi_1|\Psi_2\right>&=\left<T f_1|T f_2\right>\notag\\
  &=\sum_{s \in S}\sum_{s' \in S}f_1^*(s) f_2(s')\left<s|s'\right> \notag\\
  &=\sum_{s \in S}f_1^*(s)f_2(s)\notag \\
  &=\left<f_1|f_2\right>
\end{align}
where we used the fact $\left<s|s'\right>=\delta_{ss'}$. 

So indeed $T$ is an isomorphic transformation. And then $\mathcal{H}_I$ are equivalent to $\mathcal{H}_S$. Techniques dealing with one Hilbert Space can be directly applied to deal with the other one.

As a result of equivalence between $\mathcal{H}_I$ and $\mathcal{H}_S$, a target function $F\in \mathcal{H}_I$ for any supervised learning problem in computer vision can be encoded into a golden quantum state:
\begin{equation}
  \ket{\Psi_{G}}=\sum_{s\in S}F(s)\ket{s}
\end{equation}

\end{document}